\documentclass[preprint,12pt,3p]{elsarticle}
\usepackage{lineno}
\modulolinenumbers[5]
\journal{Neural Computing and Applications}

\usepackage{verbatim}
\usepackage{amsmath,amsthm,amsfonts,amssymb}
\usepackage{algorithm}
\usepackage{array}
\usepackage{enumerate}
\usepackage{graphics}
\usepackage{natbib}
\usepackage{algpseudocode}
\usepackage{hyperref}
\usepackage{mathrsfs}
\usepackage{subfigure}
\usepackage{multicol}
\usepackage{multirow}
\newcolumntype{C}[1]{>{\centering\let\newline\\\arraybackslash\hspace{0pt}}m{#1}}
\def\BibTeX{{\rm B\kern-.05em{\sc i\kern-.025em b}\kern-.08em
    T\kern-.1667em\lower.7ex\hbox{E}\kern-.125emX}}
\begin{document}

\begin{frontmatter}

\title{Smart Parking Space Detection under Hazy conditions  using  Convolutional Neural Networks: A Novel Approach}

\author{Gaurav  Satyanath$^{a}$\footnote{Corresponding author}, Jajati Keshari Sahoo$^{b}$, Rajendra Kumar Roul$^{c}$}
\address{$^a$Department of Electrical $\&$ Computer Engineering, Carnegie Mellon University, Pittsburgh, Pennsylvania, USA, gsatyana@andrew.cmu.edu.\\$^b$Department of Mathematics, BITS Pilani K K Birla Goa Campus, Goa, India, jksahoo@goa.bits-pilani.ac.in\\$^c$Department of Computer Science $\&$ Engineering, Thapar Institute of Engineering $\&$ Technology, Patiala, Punjab, India, raj.roul@thapar.edu}

\begin{abstract}
Limited urban parking space combined with urbanization has necessitated the development of smart parking systems that can communicate the availability of parking slots to the end users. Towards this, various deep learning based solutions using convolutional neural networks have been proposed for parking space occupation detection. Though these approaches are robust to partial obstructions and lighting conditions,  their performance is found to degrade in the presence of haze conditions. Looking in this direction, this paper investigates the use of dehazing networks that improves the performance of parking space occupancy classifier under hazy conditions. Additionally, training procedures are proposed for dehazing networks to maximize the performance of the system on both hazy and non-hazy conditions. The proposed system is deployable as part of existing smart parking systems where limited number of cameras are used to monitor hundreds of parking spaces. To validate our approach, we have developed a custom hazy parking system dataset from real-world task-driven test set of RESIDE-$\beta$ dataset. The proposed approach is tested against existing state-of-the-art parking space detectors on CNRPark-EXT  and hazy parking system datasets. Experimental results indicate that there is a significant accuracy improvement of the proposed approach on the hazy parking system dataset.
\end{abstract}


\end{frontmatter}

\section{Introduction}
The increase in population coupled with rapid urbanization has led to increased in vehicle footprints in cities.  With the urban space being limited, this has put a lot of strain on existing parking systems leading to an increase in time spent by a person searching for a parking slot. According to Giuffre et al. \cite{b1}, and Shoup et al.\cite{b2},  around $25\%$ to  $40\%$ of urban traffic flow are searching for an effective parking slot. Further, a survey conducted by Lin et al.\cite{b3} indicates that the drivers searching for a parking slot have an increased possibility for accidents due to their lack of attention to the road. This has necessitated the development of smart parking solutions which could communicate the availability of parking slots in advance to the end user, thereby reducing the time spent on road searching for a  vacant parking slot. Further, the current global market value of parking industry exceeding 2 billion US dollars has also fuelled the development of smart parking solutions. The parking industry is projected to have a compounded growth of $17\%$ over the next decade, of which $65\%$ is projected to be from smart parking solutions\footnote{Smart Parking Market worth 5.25 Billion USD by 2021. https://www.marketsandmarkets.com/ PressReleases/smart-parking.asp}.  Many research works have been done in this domain \cite{nc200, wu2018dynamic,nc1,nc8, nc6, nc9,article}.

Various existing smart parking solutions have used sensors to determine the occupancy of parking slots and communicate the same to the end user using Bluetooth/WiFi modules \cite{b5,cueva2015fuzzy,nc13, b6,b7, nc5, nc7}.  However, these solutions require installation and constant maintenance of sensors at each parking slot, making it difficult to scale to a larger IoT setting. Vision-based parking slot occupation detection offers an alternative cost-effective and scalable solution where a limited number of cameras is used to monitor hundreds of parking spaces \cite{b8,b9,b10,b11, nc20}.  However, obstruction of parking slots (for example, from trees) and changing weather conditions degrade the performance of such vision-based techniques. This problem was surpassed in \cite{b10,b11} by exploiting the robustness of Deep Neural Networks (DNN). They developed a Convolutional Neural Network (CNN) based parking space occupancy classifier for that purpose.

Despite these fine efforts, vacant parking slot detection under hazy conditions is still an open problem. In densely populated areas and areas closer to industries, the atmosphere is polluted with smoke, dust, and other particles that drastically reduce the visibility. Further, during snowfall and rainfall, the visibility is compromised. Hence, detecting parking slot occupancy under hazy conditions is challenging. All the above-mentioned vision-based approaches are not generalizable to such hazy conditions, and their performance degrades in such situations to a large extent.

Working in this direction, the proposed work focuses on improving the accuracy of vision-based parking space occupation detection  under both hazy and non-hazy conditions. To the best of our knowledge, the proposed approach is the first work that tackles the problem of parking slot occupation detection under hazy conditions.
\subsection{Contributions}
The significant contributions of this paper include the following:
\begin{enumerate}
\item[i.] A vision-based parking slot occupancy detection system is proposed that consists of the following two networks in series: an end-to-end dehazing network and a parking slot classifier (CNN). For the dehazing network, we follow \textit{All-in-One Dehazing network} ($AOD\mbox{-}Net$) architecture \cite{b23} owing to its lower computational cost than other state-of-the-art dehazing networks. For the parking space occupancy classifier, we follow the $mAlexnet$ architecture \cite{b10} owing to its low computational cost and robustness of the architecture to changing weather conditions. The proposed system is robust to partial occlusions, changing weather conditions, and the presence of haze in the image. Further, the system is deployable as part of existing smart parking systems and scalable to IoT settings.
\item[ii.] Various training procedures are explored to maximize the accuracy of the system under both hazy and non-hazy conditions. These include:
\begin{enumerate}
\item Inclusion of non-hazy images as part of the training of dehazing network.
\item Use of modified loss function during training of dehazing network incorporating a new hyperparameter for tuning the relative performance of the system on hazy and non-hazy images.
\item Joint optimization of dehazing network and classifier.
\end{enumerate}
\item[iii.] The proposed approach introduces a custom hazy parking system dataset consisting of $5010$ real world hazy occupied and unoccupied parking patches extracted from RTTS (Real-world Task-driven Test Set) subset of RESIDE-$\beta$ dataset \cite{b12}. Optional non-overlapping training and test split are also provided for benchmarking purposes. The dataset includes patches captured under various intensities of haze as well as different types of hazy conditions (e.g., snow, fog, etc.), and it will help the researchers working in this domain.
\end{enumerate}

Experimental results show that the use of dehazing network significantly improves the parking space classification accuracy (around $10-15\%$) on the proposed hazy parking system dataset. The rest of the paper includes the following: Section 2 focused on the related works in this domain. The proposed approach is discussed in Section 3, and the experimental work is carried out in Section 4. Finally, the paper is concluded in Section 5 with some future enhancements.

\section{Related work}
\subsection{Image Dehazing}

\subsubsection{Atmospheric scattering model}
The atmospheric scattering model proposed in \cite{b16,b17,b18,b19} for hazy image generation is given by
\begin{equation}\label{eq:1}
    I(x) = J(x)t(x) + A(1-t(x)).
\end{equation}
Here $A$ denotes the global atmospheric light, $J(x)$ is the clear image to be recovered, $I(x)$ is the observed hazy image,   and $t(x)$ is the transmission matrix given by
\begin{equation}\label{eq:2}
    t(x) = e^{-{\beta}d(x)}
\end{equation}
where  $\beta$ denotes the scattering coefficient of atmosphere and $d(x)$ denotes the distance between the camera and the object.

Dehazing methods involve estimating the transmission matrix $t(x)$ using physically grounded information or data-driven methods and the global atmospheric light $A$ using empirical methods. The clear image is then computed using equation \eqref{eq:1}.
\subsubsection{Dehazing networks}
Much effort has been used  to calculate the transmission matrix $t(x)$ accurately. He et al.\cite{b19} calculated the transmission matrix by discovering the dark channel prior (DCP) of the image. Zhu et al.\cite{b20} estimated the transmission map using a linear model to map the scene depth of the hazy image whose parameters are learned in a supervised way using colour attenuation prior.

Recently CNN’s are being used for haze removal. Ren et al.\cite{b21} proposed a Multi-Scale-Convolutional-Neural-Network (MSCNN) which takes a hazy image as input, outputs a coarse-scale transmission matrix which is then refined by the second fine tuning network. In \cite{b22} a dehazing network called Dehazenet is proposed, which takes a hazy image as input and outputs the corresponding transmission matrix. Both these approaches then calculate the global atmospheric light $A$ using empirical methods and the clear image using the atmospheric scattering model. However, in all the above mentioned approaches, the errors in the estimation of transmission matrix $t(x)$ and global atmospheric light $A$ will accumulate and amplify each other leading to sub-optimal results \cite{b23}.

To minimize the errors during restoration, recent works \cite{b23,b30,b31} are focused on developing an end-to-end dehazing solution. Li et al.[23] developed a novel end-to-end dehazing network based on CNN called $AOD\mbox{-}Net$. They  studied its use for high-level vision tasks such as object detection and recommended joint optimization of the pretrained dehazing network and the object detector for better performance on hazy images. Similarly, Liu et al.[31] also proposed an end-to-end CNN architecture called $``GridDehazenet"$ for single image dehazing. Working in the same direction, Ren et al.[30] suggested an end-to-end encoder-decoder based CNN called \textit{``Gated Fusion Network (GFN)"} for single image dehazing. Though $GridDehazeNet$ \cite{b31} and $GFN$ \cite{b30} achieve better dehazing results than $AOD\mbox{-}Net$, they are computationally more expensive. This complicates the implementation of $GridDehazenet$ and $GFN$ at the edges of the system for real-time applications.

\subsection{Smart Parking System}

In the early days, smart parking systems were built by embedding sensors at each parking spot. Mustaffa et al. \cite{b5} used ultrasonic sensors fixed at the ceiling of each parking spot and displayed the availability at the entrance gate. They displayed directional signage to assist the drivers in finding a parking slot. Idris et al.\cite{b6} has suggested the use of ultrasonic sensors at each parking slot and used a wireless sensor network to transmit this information to the entry gate. Khanna et al. \cite{b7} used infrared sensors at the parking slots, and the occupancy information is stored in the cloud server that is made available to the users through mobile applications. However, these approaches require installation and maintenance of sensors at each parking slot incurring huge costs, especially when the number of parking slots is large.

Vision-based techniques offer a scalable and cost-effective solution as it requires the use of only a few cameras for monitoring the entire parking system. De Almeida et al.\cite{b8} developed a new parking system dataset called PKLot consisting of around $700,000$ parking space images captured from three different cameras. The authors also proposed two systems for vacancy verification using textual descriptors: one using Local Binary Pattern (LBP) and the other using Local Phase Quantization (LPQ) \cite{b37}. Both the systems used SVM classifier and achieved an error rate of $11\%$ on the test set. The cameras are placed at the top of the building to solve the occlusion problems.

True\cite{b9} used colour histogram and Difference-of-Gaussian (DoG) using SVM and Linear Discriminant Analysis (LDA) for classification. He achieved an average accuracy of $96.45\%$ on images which are not used during the training phase. However, the problem of partial occlusion due to trees, snow, fog, etc. is not addressed in his work.

With the popularization of deep learning, various Deep Neural Networks (DNN’s) have also been proposed for parking space occupation classification \cite{nc14}. Among deep learning approaches, Convolutional Neural Networks (CNN’s) are found to be most effective for vision-based detection, and segmentation tasks \cite{b13,b14,b15}. In \cite{b10} a parking system dataset called $CNRPark$ consisting of $12,584$ parking space patches captured from 2 different viewpoints was introduced. Besides introducing the dataset, they also proposed the use of a mini version of AlexNet \cite{b13} architecture for parking space detection and proved it to be robust to light changes, partial occlusions.

Amato et al. \cite{b11} further extended the $CNRPark$ dataset to $CNRPark\mbox{-}EXT$ dataset consisting of $144,965$ parking space patches captured from 9 different cameras under various weather conditions. Using the $CNRPark\mbox{-}EXT$ dataset, they proved the robustness of  $mAlexnet$ architecture to changing camera views as well changing weather conditions. Bura et al. \cite{b36} developed a smart parking system where the license plate of the cars was captured at the entrance and used wide eye cameras for monitoring the multiple parking slots. For the real time occupancy classification, they developed a computationally inexpensive CNN architecture. More survey works in this domain can be found in \cite{nc10, nc11, nc12}.

All the above mentioned approaches are best suitable for non-hazy conditions, and their performances  are severely degraded in the presence of haze. The proposed parking system can overcome such limitations and works well both in hazy and non-hazy conditions. Experimental results show the efficiency of the proposed approach over the conventional approaches.

\section{Proposed system}
\subsection{System Overview}
The proposed parking system enhances the model proposed in \cite{b10} and is further extended to tackle hazy conditions by incorporating a dehazing network into the system’s flow. The model proposed in \cite{b10} consists of a set of cameras to monitor the entire parking lot. Periodically, images are captured from the cameras for determining the occupancy of parking spaces. From each image, all the parking spaces captured by that camera are extracted using manually created predetermined masks.  Each parking space is then classified using the trained CNN to determine its occupancy.

 The dehazing network can be employed in two different ways. In the first case, the images captured from the cameras are fed to the dehazing network. The dehazed output is then segmented into individual parking spaces and classified using the CNN. In the second case, the images captured from cameras are first segmented into parking spaces, and then each parking space is fed to the dehazing network followed by the classification process using CNN. Computationally, the former case is more efficient as it uses dehazing network once for all the parking spaces in the image. In contrast, the latter case uses dehazing network for each of the parking spaces. However, the latter case enables to jointly optimize the dehazing network and classifier as a single pipeline which can be used to achieve higher accuracies in hazy conditions.

The choice is dependent on the computational resources available, the number of parking spaces in the parking lot, and the period between successive capture of images by the cameras. In this paper, the second case is adopted for experimental works. Accordingly, the classifier and dehazing network are developed considering the computational feasibility of the system.

\begin{figure}
\begin{center}
\includegraphics[scale=0.9]{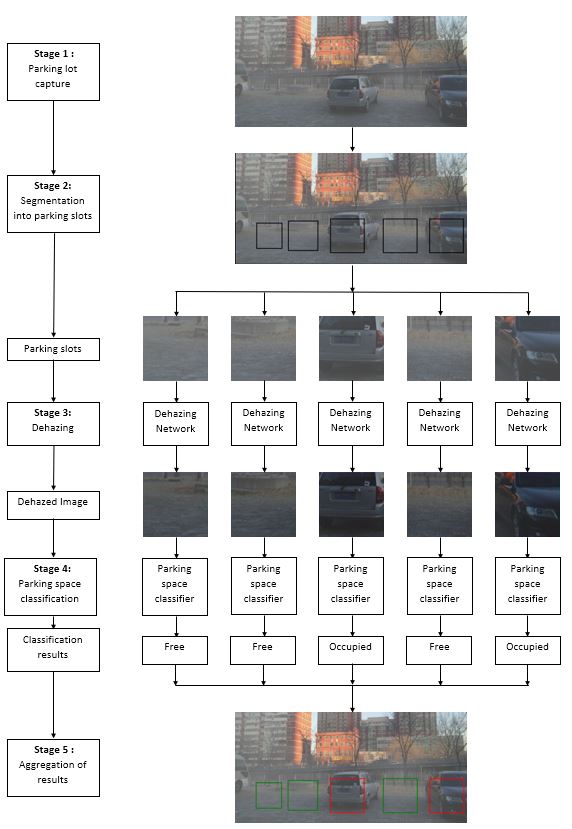}
\caption{Overview of the proposed approach.}
\label{fig:2}
\end{center}
\end{figure}

The proposed system consists of five stages as follows:
\begin{itemize}
\item[Stage 1]: Images of the parking lot are captured using cameras.
\item [Stage 2]: The portion of parking slots from each image are then extracted using predetermined masks. Each segmented image is resized to $224 \times 224$.
\item [Stage 3]: Each parking slot is passed through the dehazing network.
\item [Stage 4]: The parking slot occupancy is determined using  CNN.
\item[Stage 5]: Finally, all the classification results are aggregated.
\end{itemize}
Figure \ref{fig:2} demonstrates the overview of the proposed approach.

 The following sections describe how the CNN is used for the parking space classification process and how the dehazing network is used in the proposed model.

\subsection{Parking space classification}
For parking space classification, the proposed system followed the mAlexnet proposed in \cite{b10}. mAlexnet is a miniaturized version of Alexnet proposed in \cite{b13}, specifically built for binary classification tasks. It is computationally cheaper and is found to achieve accuracies similar to that of Alexnet for binary classification tasks \cite{b11}. mAlexnet consists of three convolutional layers and two fully connected layers. Each convolutional layer is followed by ReLU activation and max pooling layers. The fully connected layers FC4 and FC5 are followed by ReLU and SoftMax activations, respectively.

The mAlexnet architecture proposed in \cite{b10} uses Local response normalization. However quantitative experiments performed by Z. Wang et.al in \cite{b38} on CIFAR-10 dataset using various CNN networks showed batch normalization to be a better normalization technique than local response normalization. Taking inspiration from \cite{b38}, the proposed system uses batch normalization layer instead of local response normalization layer (LRN) after first two convolutional layers.   For conv1-3, the number and size of filter is specified by ``num×width×height+stride” along with ReLU activation and  batch normalization. Max pooling layer is denoted by ``width×height+stride”. For fully connected layers, the dimensionality along with the activation function is presented. Padding is not used for any of the layers. The details of the modified mAlexnet architecture used in the proposed system are illustrated in Table \ref{table:1}.

\begin{table}
\caption{Modified mAlexnet architecture used in the proposed approach}
\begin{tabular}{|C{3cm}|C{3cm}|C{3cm}|C{2.625cm}|C{2.625cm}|}
\hline
Conv1 & Conv2 & Conv3 & Fc4 & Fc5 \\
\hline

16x11x11+4 & 20x5x5+1 & 30x3x3+1 & & \\
pool 3x3+2 & pool 3x3+2 & pool 3x3+2 & 48 & 2 \\
Batchnorm & Batchnorm & - & ReLU & Softmax \\
ReLU & ReLU & ReLU & & \\
\hline
\end{tabular}
\label{table:1}
\end{table}

\subsection{Dehazing network}
\begin{figure}
\begin{center}
\includegraphics[scale=0.5]{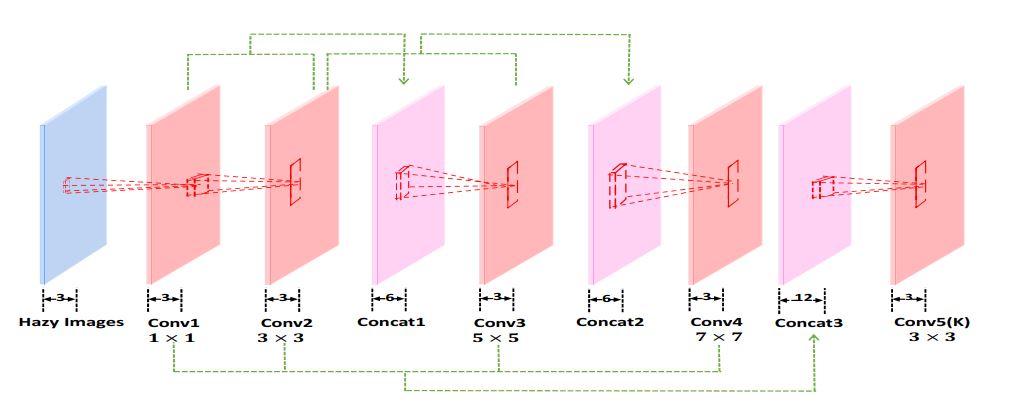}
\caption{The network architecture of K-estimation module.}
\label{fig:3}
\end{center}
\end{figure}
The proposed system followed the \textit{AOD-Net} proposed in [23].  It combines all the parameters of atmospheric scattering equation by rewriting it as follows.
\begin{align}
J(x)& =\frac{I(x)}{t(x)}-\frac{A}{t(x)}+A \tag{3} \\
& =K(x)I(x)-K(x)+b, where \tag{4} \label{eq:4} \\
K(x)& =\frac{\frac{(I(x)-A)}{t(x)}+(A-b)}{I(x)-1} \tag{5}
\end{align}

Where $b$ is constant bias with a default value of 1.   With $t(x)$ and $A$ incorporated into a single parameter $K(x)$, a CNN architecture named as the K-estimation module is trained to learn the parameter $K(x)$ given an input image $I(x)$. It consists of five convolutional layers and combines features from different layers to compensate for information loss during convolution. \textit{``concat1”} combines features of \textit{``conv1”} and \textit{``conv2”}.\textit{``concat2”} combines features of \textit{“conv1”,``conv2”} and \textit{``conv3”. ``concat3”} combines features of \textit{``conv1”,``conv2”, ``conv3”} and \textit{``conv4”}. Each convolutional layer is followed by a ReLU activation layer. The CNN architecture is depicted in Figure \ref{fig:3}\cite{b23}. Once $K(x)$ is estimated, the clear image is obtained using equation \ref{eq:4}.

Experimentally, AOD-Net[23] is found to have lower average runtime than other dehazing networks \cite{b31}. Given the computational constraints of the system, AOD-Net is the ideal choice when compared to other works such as GridDehazenet \cite{b31}, Gated Fusion Network \cite{b30}, MSCNN \cite{b21}, and Dehazenet \cite{b22} owing to its small network architecture.

\section{Experimental Analysis}
\subsection{Datasets used for evaluation}
\subsubsection{Parking lot datasets}
For parking space occupancy classification, the proposed system used CNRPark\cite{b10}, CNRPark-EXT\cite{b11} and PKLot\cite{b8} datasets. PKLot consists of around $700,000$ manually checked and labeled parking space images acquired from parking lots of Federal University of Parana (UFPR) and Pontifical Catholic University of Parana (PUCPR) located in Curitiba, Brazil. The dataset consists of images captured from different parking lots with distinct features and is divided into three subsets: UFPR04, UFPR05, and PUCPR corresponding to images captured by cameras at $4^{th}$ and $5^{th}$ floor of UFPR, and at $10^{th}$ floor of PUCPR respectively. With the cameras placed at the top of the buildings, the dataset contains fewer occluded images. The dataset also contains images under different weather conditions (Sunny, Overcast, Rainy) and various illuminations occasioned by weather conditions. The dataset contains both non-segmented full images of the parking lot and pre-segmented parking patches. The slope of pre-segmented parking space rectangle is rotated such that those with a slope between $0^{\circ}$ to $45^{\circ}$ were rotated to $0^{\circ}$ and those with a slope between $45^{\circ}$ to $90^{\circ}$ were rotated to $90^{\circ}$. Details of the PKLot dataset are shown in Table \ref{table:2}.

The CNRPark (CNRPark A and CNRPark B) dataset consists of $12,584$ images segregated into two non overlapping subsets: CNRPark A (Images captured using camera A) consisting of $6,171$ images and CNRPark B (Images captured using camera B) consisting of $6,413$ images. The dataset includes patches captured under various light conditions. It also contains patches partially occluded by trees as well as shadowed by the neighbouring cars. This enables us to demonstrate the robustness of the classifier to changing real life conditions. Further, one can also test the robustness of the classifier to changing camera views by training on images captured from one camera and testing on images captured from different cameras. CNRPark A contains fewer occluded images than CNRPark B.

CNRPark dataset is extended to CNRPark-EXT in \cite{b11}. CNRPark-EXT consists of $144,965$ image patches captured from $9$ different cameras. Those patches are captured under various weather conditions (Sunny, Overcast, Rainy) as well as at different distances from the camera. The dataset also contains patches partially occluded or shadowed, enabling to train the classifier on various difficult real-life scenarios. The dataset contains both non-segmented full images of the parking lot and pre-segmented parking patches. Further, different training, validation and test splits are provided to enable common grounds for testing the classification algorithms. Details of the CNRPark and CNRPark-EXT datasets used in the experiments are shown in Table \ref{table:2}.

\begin{table}
\begin{center}
\caption{Details of CNRPark, CNRPark-EXT and PKLot datasets \cite{b8},\cite{b11},\cite{b10}.}
\begin{tabular}{cccc}
\hline
Subsets & Free patches & Busy patches & Total patches \\
\hline

CNRPark A & 2549 & 3622 & 6171 \\
CNRPark B & 1632 & 4781 & 6413 \\
CNRPark & 4181 & 8403 & 12584 \\ \\

CNRPark-EXT	& 65658	& 79307 & 144965 \\
CNRPark-EXT TEST & 13549 & 18276 & 31825 \\
CNRPark + EXT Train C1-C8 & 16784 & 21769 & 38553 \\
CNRPark + EXT Train	& 51059 & 56018	& 107077 \\ \\
PKLot Train & 27314 & 41744 & 69058 \\
PKLot Test & 275894 & 248583 & 524477 \\
\hline
\end{tabular}
\label{table:2}
\end{center}
\end{table}

\subsubsection{Single Image Dehazing datasets}
For training the dehazing network, the proposed system used  Outdoor Training Set (OTS) of Reside-$\beta$ dataset \cite{b12}. OTS dataset consists $72,135$ synthetic outdoor hazy images synthesized from 2061 diverse non-hazy outdoor images\footnote{``Beijing realtime weather photos," http://goo.gl/svzxLm.}. Each non-hazy image was used to synthesize $35$ synthetic hazy images by using different pairs of $A$ and $\beta$ of atmospheric scattering model described in equation \ref{eq:1}. Values of $A \in [0.8\mbox{-}1]$ with increment of 0.05 and $\beta \in [0.04, 0.06, 0.08, 0.1, 0.12, 0.16, 0.2 ]$. This captures various hazy conditions encountered in real life. The depth map for each image was estimated using the procedure outlined in \cite{b25}. To counter the errors during depth estimation, the final hazy images were visually inspected for any irregularities.

\subsubsection{Hazy Parking System dataset}

The proposed system creates a benchmark for the hazy parking system using Real-time Task-driven Test Set (RTTS) of RESIDE-$\beta$ dataset \cite{b12}. RTTS consists of $4322$ images extracted from the Web covering real world hazy traffic and driving scenarios.  Each image is annotated with bounding boxes and object categories.  Hazy car images and hazy unoccupied parking space patches are manually extracted to create the hazy parking system dataset. Optional non-overlapping training and test splits are provided for the sake of comparing classification results under hazy conditions. \\

\begin{figure*}[!tb]
\includegraphics[width=2.5cm, height=2.5cm]{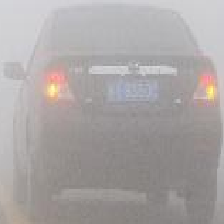} \hspace{0.7cm}
\includegraphics[width=2.5cm, height=2.5cm]{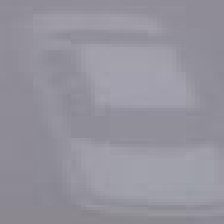} \hspace{0.7cm}
\includegraphics[width=2.5cm, height=2.5cm]{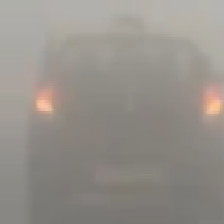} \hspace{0.7cm}
\includegraphics[width=2.5cm, height=2.5cm]{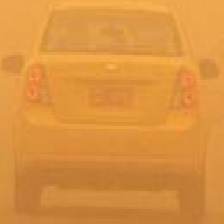} \hspace{0.7cm}
\includegraphics[width=2.5cm, height=2.5cm]{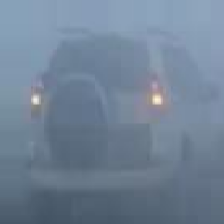} \\ \\
\includegraphics[width=2.5cm, height=2.5cm]{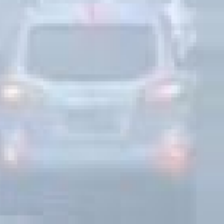} \hspace{0.7cm}
\includegraphics[width=2.5cm, height=2.5cm]{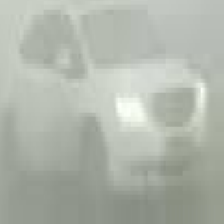} \hspace{0.7cm}
\includegraphics[width=2.5cm, height=2.5cm]{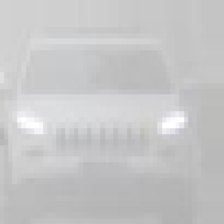} \hspace{0.7cm}
\includegraphics[width=2.5cm, height=2.5cm]{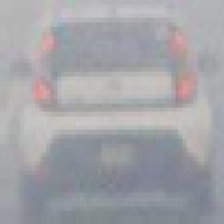} \hspace{0.7cm}
\includegraphics[width=2.5cm, height=2.5cm]{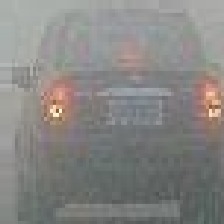} \\ \\
\includegraphics[width=2.5cm, height=2.5cm]{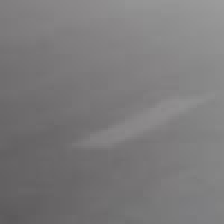} \hspace{0.7cm}
\includegraphics[width=2.5cm, height=2.5cm]{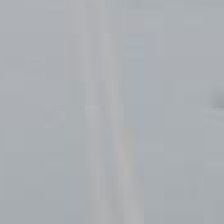} \hspace{0.7cm}
\includegraphics[width=2.5cm, height=2.5cm]{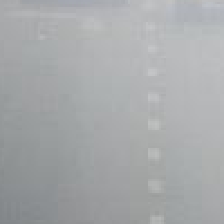} \hspace{0.7cm}
\includegraphics[width=2.5cm, height=2.5cm]{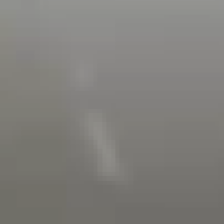} \hspace{0.7cm}
\includegraphics[width=2.5cm, height=2.5cm]{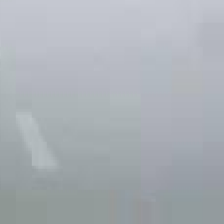} \\ \\
\includegraphics[width=2.5cm, height=2.5cm]{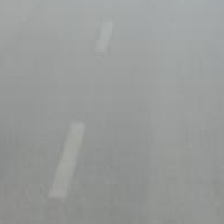} \hspace{0.7cm}
\includegraphics[width=2.5cm, height=2.5cm]{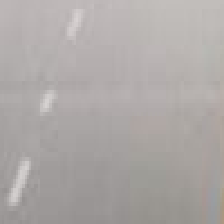} \hspace{0.7cm}
\includegraphics[width=2.5cm, height=2.5cm]{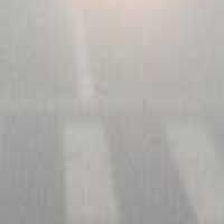} \hspace{0.7cm}
\includegraphics[width=2.5cm, height=2.5cm]{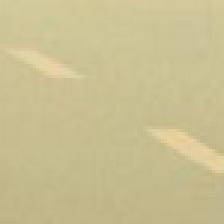} \hspace{0.7cm}
\includegraphics[width=2.5cm, height=2.5cm]{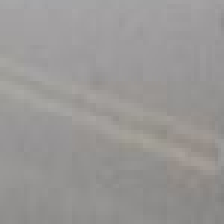}
\caption{Sample Images of the proposed Hazy Parking system Dataset: Top two rows - Hazy busy parking patches, Bottom two row - Hazy unoccupied parking patches}
\label{fig:4}
\end{figure*}

The dataset contains $752$ unaugmented patches with $502$ hazy car patches, and $250$ hazy free patches. It is splited into non-overlapping $70\%$ training and $30\%$ test set. The test set contains roughly the same number of busy and free patches before augmentation. The images in the training and test set are then augmented using Keras\footnote{Chollet, Franccois et al. ``Keras." https://keras.io. (2015).}(a deep learning framework) image data pre-processing tool. The hazy car patches are randomly flipped horizontally and a crop of the image containing $90\%$ of the total height and width is considered. Hazy unoccupied patches are also augmented in a similar setting where additionally the patches are  randomly flipped in a vertical manner. The dataset consists of $5010$ augmented labelled patches, and captures busy and unoccupied patches under various hazy conditions such as fog, snow and pollution with varying intensity levels from lightly hazed to heavily hazed patches. Details of the dataset is shown in Table \ref{table:3}. For demonstration purposes, we have shown some of the sample images of the proposed hazy parking system dataset in Figure \ref{fig:4}. Out of 5010 images, 500 images (250 of training set and 250 of test set) of the hazy parking system dataset is made available publicly for download\footnote{https://github.com/GauravS9776/Hazy-parking-system}.

\begin{table}[h]
\begin{center}
\caption{Details of Hazy parking system dataset.}
\begin{tabular}{cccc}
\hline
Datasets & Free patches & Busy patches & Total patches \\
\hline
Unaugmented Train & 150 & 380 & 532 \\
Unaugmented Test & 100 & 122 & 222 \\
Unaugmented images & 250 & 502 & 752 \\ \\
Hazy parking system Train	& 1500 & 1900 & 3400 \\
Hazy parking system Test & 1000 & 610 & 1610 \\
Hazy parking system & 2500 & 2510 & 5010 \\
\hline
\end{tabular}
\label{table:3}
\end{center}
\end{table}

\subsection{Parking slot classifier}\label{4.2}

Amato et al.\cite{b10} achieved state-of-the-art results using the mAlexnet CNN architecture. The mAlexnet network takes an input image of size $224 \times 224$. They trained mAlexnet on Caffe framework \cite{b32} using gradient descent with momentum. Momentum and weight decay were set to $0.9$ and $5\times10^{-4}$  respectively. The learning rate was individually determined for each experiment and decreased by a factor of 10 after loss stabilization. All the models were trained for 30 epochs. Further in \cite{b11}, mAlexnet's performance was proven to be comparable to that of Alexnet\cite{b13} on CNRPark \cite{b10} and CNRPark-EXT\cite{b11} datasets while being three times smaller in size. In \cite{b11}, all the models were trained using gradient descent with momentum for 6 epochs, having an initial learning rate of $0.0008$, which was multiplied by $0.75$ after every 2 epochs. The batch size was set to $64$, momentum to $0.9$, and  weight decay to $0.0005$.

In this work, the proposed system uses the mAlexnet architecture after doing certain modifications to it (shown in Table \ref{table:1}). Training and testing images are shuffled and resized to 224×224 pixels. The modified mAlexnet architecture is trained on Caffe framework using Adam optimization algorithm\footnote{https://machinelearningmastery.com/adam-optimization-algorithm-for-deep-learning/} with the following hyperparameters: $\beta_{1}=0.9$, $\beta_{2}=0.999$, and weight decay=$5\times10^{-4}$. The learning rate is fixed to $0.001$ throughout the training process. The training procedure followed in this work is shown in Table \ref{table:4}.

The proposed trained model is compared with the existing mAlexnet model of \cite{b10} and \cite{b11} using CNRPark, CNRPark-EXT and PKLot datasets (shown in Tables \ref{table:5} and \ref{table:6}). For comparison, we trained the models on various subsets of CNRPark, CNRPark-EXT, PKLot datasets and are evaluated on non-overlapping test sets. Models shown in Tables \ref{table:5} and \ref{table:6} are trained for 10 epochs and 5 epochs respectively. All the models were trained on Ubuntu $18.04$ virtual environment with 16 GB RAM running on Intel(R) Core(TM) i5-7300HQ CPU @ 2.50GHz in CPU mode.

It can be observed from Tables \ref{table:5} and \ref{table:6} that the proposed model using modified mAlexnet architecture achieves better results compare to the existing Alexnet and mAlexnet pretrained models of \cite{b10} and \cite{b11} on subsets of CNRPark, CNRPark-EXT, and PKLot datasets.  This can be attributed to two factors: Use of batch normalization instead of local response normalization and use of Adam optimization over gradient descent with momentum for training.

\begin{table}
\caption{Training procedure  for modified mAlexnet}
\begin{tabular}{C{2.75cm}C{4.5cm}C{2.5cm}C{2.5cm}C{2cm}}
\hline
Optimization Algorithm & Hyperparameters & Initial Learning rate & Batch size & Epochs \\
\hline
Adam & Weight decay: 5×10$^{-4}$ $\beta$1=0.9, $\beta$2=0.999 & 0.001 & 64 & 5-10 \\
\hline
\end{tabular}
\label{table:4}
\end{table}

\begin{table}
\begin{center}
\caption{Comparison of accuracy of proposed model with pretrained models of \cite{b10}}
\begin{tabular}{ccccc}
\hline
Train & Test & Network & Base learning rate & Accuracy \\
\hline
\multirow{2}{*}{CNRPark A} & \multirow{2}{*}{CNRPark B} & modified mAlexnet & 0.001 & \textbf{87.56\%} \\
& & mAlexnet\cite{b10} & 0.001 & 86.30\% \\
\hline
\multirow{2}{*}{CNRPark B} & \multirow{2}{*}{CNRPark A} & modified mAlexnet & 0.001 & \textbf{90.78\%} \\
& & mAlexnet\cite{b10} & 0.001 & 90.70\% \\
\hline
\end{tabular}
\label{table:5}
\end{center}
\end{table}

\begin{table}
\begin{center}
\caption{Accuracy comparison of proposed model with the pretrained models of \cite{b11}}
\begin{tabular}{cccc}
\hline
Train set & Test set & Network & Accuracy \\
\hline
 & & modified mAlexnet & \textbf{93.92\%} \\
CNRPark All & CNRPark-EXT Test & mAlexnet\cite{b11} & 93.70\% \\
& & Alexnet\cite{b11} & 93.38\% \\
\hline
 & & modified mAlexnet & \textbf{97.85\%} \\
CNRPark+EXT TRAIN C1-C8 & CNRPark-EXT Test & mAlexnet\cite{b11} & 96.44\% \\
 & & Alexnet\cite{b11} & 96.74\% \\
\hline
 & & modified mAlexnet & \textbf{98.51\%} \\
CNRPark+EXT TRAIN & CNRPark-EXT Test & mAlexnet\cite{b11} & 97.78\% \\
& & Alexnet\cite{b11} & 97.98\% \\
\hline
& & modified mAlexnet & \textbf{98.89\%} \\
PKLot Train & PKLot Test & mAlexnet\cite{b11} & 97.96\% \\
& & Alexnet\cite{b11} & 98.76\%\\
\hline
\end{tabular}
\label{table:6}
\end{center}
\end{table}

\subsection{Dehazing network}
\subsubsection{Model description}
For the dehazing network, the proposed model followed the AOD-Net\cite{b23} architecture. Instead of evaluating the dehazing networks on performance metrics such as peak signal-to-noise ratio (PSNR) and structural similarity index (SSIM), they are evaluated by their utility before the parking space classification takes place on hazy and non-hazy datasets.  By using dehazing network before the classification process starts will improves the performance of the system on hazy images, but degrades the performance on non-hazy images. Alternative training methodologies for dehazing networks are explored to improve the performance of system on non-hazy images.  For that purpose, we explore four different models, and the details are presented below:
\begin{enumerate}
\item \textit{Model 1}: AOD-Net trained on hazy images of OTS dataset followed by parking space classifier.
\item \textit{Model 2}: AOD-Net trained on both hazy and clear (non-hazy) images of OTS dataset followed by parking space classifier.
\item \textit{Model 3} : AOD-Net trained on both hazy and clear (non-hazy) images of OTS dataset using a modified loss function $L_1$ followed by parking space classifier. The modified loss function $L_1$ is given by:
\[L_1=y{\lambda}L+(1-y)(1-{\lambda})L \tag{6} \label{eq:6} \]
Where $L$ denotes the Mean Square Error (MSE) loss function between the output of the network and the ground truth of the image corresponding to the input. $y=1$, when the input is a clear (non-hazy) image, and $y=0$, when input is a hazy image. $\lambda$ is a hyper-parameter whose value ranges from $0$ to $1$. $\lambda$ can be used to tune the relative performance of the network on hazy or non-hazy images.  For $0\leq\lambda<0.5$, the network is trained to perform better on hazy images. For $0.5<\lambda\leq1$, the network is trained to perform better on non-hazy images. This gives more flexibility than other three models to train the system according to specific requirements. For experiments, four values of $\lambda$  are used to train four networks, each randomly chosen from the following four intervals: $[0,0.24],[0.25,0.49],[0.5,0.74],[0.75,1]$.
\end{enumerate}
We then compare the performance of the above models with modified mAlexnet, mAlexnet \cite{b10}, and Alexnet \cite{b11} trained on CNRPark-EXT train set.

The combined dehazing network and classification process necessitates the output of AOD-Net, and input of the modified mAlexnet classifier to have same dimensions. The classifier developed in the section \ref {4.2} takes the inputs of 224×224×3 images. Further, as the input and output of AOD-Net have same dimensions, the training and testing images are cropped to $224\times224$.

\subsubsection{Model Training}
AOD-Net is trained using gradient descent with momentum having the following hyperparameters: momentum = $0.9$ and weight decay = $0.0001$. For non-hazy images, Mean Square Error (MSE) between the output of the network and the ground truth of the image is used as the loss function. For clear (non-hazy) images, the output of the dehazing network is compared with the input for loss calculation. The number of clear (non-hazy) images used during the training phase is kept low compared to that of hazy images to prevent the network from learning an identity function between the input and output. The gradients are constrained within $[-0.1,0.1]$ to stabilise the training. The model is trained using Caffe framework with a fixed learning rate of $0.001$. AOD-Net takes around 4-5 epochs to converge on the OTS dataset. Hence, all the networks are trained for 5 epochs. The same procedure is adopted for all the models except Model 3, where the loss function is altered as given in equation \ref{eq:6}. The dehazing network is trained on Nvidia K80 GPU provided as part of p2.xlarge EC2 instance by Amazon Web Services (AWS) using Jetware’s `Caffe Python 2.7 Nvidia GPU production on Ubuntu’ AMI. The training procedure followed is summarized in Table \ref{table:7}.

It is found that joint tuning of pretrained AOD-Net and modified mAlexnet as a unified pipeline improves the classification performance of the system under hazy conditions. All the trained AOD-Net models are concatenated with modified mAlexnet and are jointly tuned as a single network on hazy parking system training set. The training is performed on Caffe framework using gradient descent with momentum having the following hyperparameters: momentum = $0.9$ and weight decay = $0.0001$. The model is trained with a fixed learning rate of $0.0001$ for 5 epochs, and the gradients are constrained between $[-0.1,0.1]$ to stabilise the training.

\begin{table}
\caption{Training procedure  for AOD-Net and joint optimization of AOD-Net and modified mAlexnet pipeline.}
\begin{tabular}{C{2cm}C{2.5cm}C{4.25cm}C{2.5cm}C{1.25cm}C{1.5cm}}
\hline
Network & Optimization Algorithm & Hyperparameters & Initial Learning rate & Batch size & Epochs \\
\hline
AOD-Net & Gradient Descent with momentum & Weight decay: 1×10$^{-4}$ Momentum : 0.9 & 0.001 & 64 & 5 \\
AOD-Net + modified mAlexnet & Gradient Descent with momentum & Weight decay: 1×10$^{-4}$ Momentum : 0.9 & 0.001 & 64 & 10 \\
\hline
\end{tabular}
\label{table:7}
\end{table}
\subsubsection{Discussion on Empirical Results}
All the models are evaluated using hazy parking system test set (Table \ref{table:3}) and CNRPark-EXT test set (Table \ref{table:2}) to test the model's performance under both hazy and non-hazy conditions, respectively. The accuracies of the four models on these datasets are displayed in Table \ref{table:8}. The proposed modified mAlexnet model performs extremely well under non-hazy conditions achieving $98.51\%$ accuracy on CNRPark-EXT test set. However, it performs poorly under hazy conditions achieving $74.41\%$ accuracy on hazy parking system test set. Similarly mAlexnet\cite{b10} and Alexnet\cite{b11} achieves $97.76\%$ and $97.98\%$ on CNRPark-EXT test set while achieving $76.27\%$ and $80.37\%$ on hazy parking test set respectively. In comparison, the proposed AOD-Net+mAlexnet model achieved significant gains of over $10\mbox{-}15$\%  accuracy on hazy parking system test set with a marginal decrease in accuracy of $1\mbox{-}3\%$ on CNRPark-EXT test set.

\begin{table}[h]
\caption{Comparison of model accuracies on Hazy parking system – Test set and CNRPark-EXT Test set.}
\begin{tabular}{C{8cm}C{3.6cm}C{3.675cm}}
\hline
Net & Accuracy on Hazy parking system Test & Accuracy on CNRPark-EXT Test  \\
\hline
Model 1 & 86.71\% & 97.25\% \\
Model 1 after joint optimization & 88.39\% & 96.68\% \\ \\
Model 2 & 86.40\% & 97.28\% \\
Model 2 after joint optimization & 86.89\% & 96.72\% \\ \\
Model 3 ($\lambda$=0.1916) & 86.27\% & 97.34\%; \\
Model 3 ($\lambda$=0.1916) after joint optimization & 88.07\% & 96.88\% \\ \\
Model 3 ($\lambda$=0.4291) & 86.89\% & 96.61\% \\
Model 3 ($\lambda$=0.4291) after joint optimization & 88.63\% & 96.42\% \\ \\
Model 3 ($\lambda$=0.6555) & 86.96\% & 97.38\% \\
Model 3 ($\lambda$=0.6555) after joint optimization & 86.27\% & 96.33\% \\ \\
Model 3 ($\lambda$=0.8984) & 84.78\% & 97.57\% \\
Model 3 ($\lambda$=0.8984) after joint optimization & 86.71\% & 97.08\% \\ \\
Modified mAlexnet & 74.41\% & 98.51\% \\
mAlexnet\cite{b10} & 76.27\% & 97.76\% \\
Alexnet\cite{b11} & 80.37\% & 97.98\% \\
\hline
\end{tabular}
\label{table:8}
\end{table}

Model 2 achieves higher accuracy than Model 1 on CNRPark-EXT test set, while achieving lower accuracy on hazy parking test set. However, after joint optimization, Model 1 achieves higher accuracy than Model 2 on hazy parking system test set. Both these observations can be attributed to the fact that AOD-Net in Model 2 was trained on both hazy and clear (non-hazy) images, thereby enhancing system’s performance under non-hazy conditions while degrading system’s performance under hazy conditions.

In the case of Model 3, $\lambda$ is a hyperparameter that can be used to tune the relative performance of the network on hazy and non-hazy images. In accordance with equation \ref{eq:6}, it is found that models trained with $\lambda=0.1916$ and $\lambda=0.4291$ achieve lower accuracy than models trained with $\lambda=0.6555$ and $\lambda=0.8984$ on CNRPark-EXT test set. Models trained with  $\lambda=0.1916$ and $\lambda=0.4291$ achieve higher accuracy than the model trained with $\lambda=0.8984$ on hazy parking test set and slightly lower accuracy than model trained with $\lambda=0.6555$. Further, model 3 trained with $\lambda=0.6555$ outperforms Model 1 and Model 2 under both hazy and non-hazy conditions.

Joint optimization is found to improve the system’s performance on hazy images in almost all of the cases, while it is found to degrade the system’s accuracy under non-hazy conditions in all the cases. However, in some cases (Model 1, Model 3 $\lambda=0.1916,0.4291,0.8984$), joint optimization is found to be beneficial as it is found to increase the accuracy on hazy parking test set by $1.5\%$ with around $0.5\%$ decrease in accuracy on CNRPark-EXT test set. Model 3 ($\lambda=0.4291$) after joint optimization achieves the highest accuracy of $88.63$\% on hazy parking system test set. Apart from Model 4, Model 3 ($\lambda=0.8984$) achieves the highest accuracy of $97.57\%$ on CNRPark-EXT test set. This can be attributed to $\lambda$ value used for training the dehazing network being closer to 1. Model 3 with $\lambda=0.8984$ achieves around $10\%$ gain on hazy parking system test set with only $1\%$ decrease in accuracy on CNRPark-EXT test set. Model 3 with $\lambda=0.6555$ after joint optimization achieving over $12\%$ accuracy gain on hazy parking test set, and  is particularly useful when the target application requires the system to be robust against the presence of haze in the image.

\subsection{Runtime analysis}

We compare the average runtimes of AOD-Net \cite{b23} with other state-of-the-art dehazing networks such as MSCNN\cite{b21} Dehazenet\cite{b22}, GridDehazenet\cite{b31} and GFN\cite{b30}. Table \ref{table:10} \cite{b31} shows the average runtimes of these models. AOD-Net is atleast $2.5$ times faster than the other state-of-the-art dehazing networks.
\begin{table}[h]
\caption{Average runtime of state-of-the-art dehazing networks \cite{b31}.}
\begin{tabular}{C{2.2cm}C{2cm}C{3cm}C{2.25cm}C{2.75cm}C{1.75cm}}
\hline
Network & AOD-Net\cite{b23} & Grid-DehazeNet\cite{b31}& MSCNN\cite{b21} & Dehazenet\cite{b22} & GFN\cite{b30} \\
\hline
Average runtime (s) & 0.08 & 0.22 & 0.26 & 0.30 & 0.37 \\
\hline
\end{tabular}
\label{table:10}
\end{table}

\begin{table}[h]
\begin{center}
\caption{ Average runtime of proposed system(AOD-Net and modified mAlexnet in series), modified mAlexnet , mAlexnte \cite{b10} and Alexnet\cite{b11}  architectures}
\begin{tabular}{C{1.75cm}C{3.5cm}C{2.75cm}C{3.75cm}C{2.5cm}}
\hline
Network & Average runtime on CNRPark-EXT Test (s) & Average runtime on PKLot Test (s) & Average runtime on Hazy parking system Test (s) & Mean Average runtime (s) \\
\hline
mAlexnet & 0.009 & 0.009 & 0.009 & 0.009\\ \\
modified mAlexnet & 0.009 & 0.009 & 0.009 & 0.009\\ \\
Alexnet & 0.181 & 0.182 & 0.180 & 0.181 \\ \\
AOD-Net + modified mAlexnet & 0.091 & 0.090 & 0.093 & 0.0907 \\
\hline
\end{tabular}
\label{table:9}
\end{center}
\end{table}

Table \ref{table:9} compares the average runtime of the proposed AOD-Net and mAlexnet pipeline with modified mAlexnet, Alexnet\cite{b11},mAlexnet \cite{b10}. We randomly select $1000$ random $224\times 224$ images from each CNRPark-EXT test set and PKLot test set and $200$ random $224\times224$ images from hazy parking system test set to run for each model on the same machine without GPU acceleration. The model runtime is tested using Caffe framework.

Modified mAlexnet and the existing mAlexnet with its compact nature are 20 times faster than Alexnet while achieving similar accuracies on parking space classification, making it an ideal choice for real time applications. With the use of dehazing network, the average runtime of the proposed architecture is significantly higher than mAlexnet architecture. However, the proposed pipeline of AOD-Net and mAlexnet in series is at least 2 times faster than Alexnet and other state-of-the-art dehazing networks. The runtime analysis is performed on Ubuntu 18.04 virtual environment with 16 GB RAM running on Intel(R) Core(TM) i5-7300HQ CPU @ 2.50GHz.

\section{Conclusion}
This paper proposed a parking space occupancy detection model that is robust to hazy conditions. The system pipeline consists of two parts: A dehazing network followed by the parking space classification process using CNN. The system is evaluated on both hazy and non-hazy images.  Modified mAlexnet architecture is trained to classify parking space occupancy and is compared against other state-of-the-art architectures. With modification in existing mAlexnet’s architecture and the training methodology, higher classification accuracies are achieved on CNRPark and CNRPark-EXT datasets.

For the dehazing network, we trained AOD-Net, an end-to-end dehazing network on OTS dataset, and evaluated its performance by its utility before the classification. We studied the use of clear (non-hazy) images for training the dehazing network and joint optimization of dehazing network.  Further, the trade-off between model’s performance on hazy and non-hazy images is addressed, and a new loss function $L_1$ is introduced. We incorporated a hyperparameter $\lambda$, which can be used to tune the relative performance of the model on hazy and non-hazy images. This enables us to tune the proposed system according to target requirements.

To validate our approaches on hazy images, we have created a new hazy parking system dataset consisting of hazy and unoccupied parking patches extracted from RTTS subset of Reside-$\beta$ dataset. CNRPark-EXT is used to evaluate the model’s performance on non-hazy images. Experimental results show that the proposed approach achieved significant gains over the parking space classification system on hazy images.
However, with the use of dehazing network, the average runtime of the proposed pipeline significantly greater than that of mAlexnet architecture.

This work can be extended by using quantization techniques, which will reduce the runtime of the proposed system. Further, the parameter $\lambda$  used in the training of dehazing networks can be extensively tuned for better performance.

\noindent{\bf Conflicts of interest }\\
No potential conflict of interest was reported by the authors.

\end{document}